\documentclass{arxiv} 

\begin{document}

\newcommand{\bvmyear}{2024}

\selectlanguage{english} 

\title{exploreCOSMOS }

\subtitle{Interactive Exploration of Conditional Statistical 
\\
Shape Models in the Web-Browser}

\titlerunning{exploreCOSMOS}

\author{
     Maximilian \lname{Hahn},
     Bernhard \lname{Egger}
}

\authorrunning{Hahn \& Egger}

\institute{
Cognitive Computer Vision Lab, Department of Computer Science
Friedrich-Alexander-Universtitat Erlangen-Nürnberg
}

\email{maximilian.hahn@fau.de, bernhard.egger@fau.de}

\maketitle
\begin{abstract}
Statistical Shape Models of faces and various body parts are heavily used in medical image analysis, computer vision and visualization. Whilst the field is well explored with many existing tools, all of them aim at experts, which limits their applicability. We demonstrate the first tool that enables the convenient exploration of statistical shape models in the browser, with the capability to manipulate the faces in a targeted manner. This manipulation is performed via a posterior model given partial observations.
We release our code and application on GitHub \url{https://github.com/maximilian-hahn/exploreCOSMOS}.
\end{abstract}

\section{Introduction}
Statistical shape models (SSMs) serve as a powerful tool for analyzing and interpreting complex shapes in various fields, such as medical imaging, computer graphics and computer vision~\cite{luthi2017gaussian}.
One key motivation for utilizing SSMs is their ability to overcome the limitations of individual shape analysis. Traditional methods often struggle to effectively capture the variations found within a shape dataset, making it challenging to accurately represent and compare shapes. SSMs address this issue by aligning the set of shapes to a reference shape and thus being able to create a mean shape and generate a comprehensive representation of shape variation.
We propose an interactive web-based application that allows the user to visualize and modify instances of their SSM via a 3D mesh. Shape variations of the model can be explored and the user can add constraints such that a posterior model can be computed that alters the appearance of the model by satisfying those user-given constraints. These constraints could be the movement of selected vertices of the 3D mesh representing the model or marking them with either predefined or user-created landmarks, ensuring that these vertices remain in place. With these constraints, the posterior mesh has vertices at the defined positions, and the statistically most probable positions for the rest of the vertices are computed so that the whole shape fits as well as possible into the statistical model. This enables the user to explore a SSM, aid in avatar creation or 3D modeling tasks, and realize visualizations by generating faces with desired features~\cite{albrecht2013posterior, luthi2017gaussian, basso2005statistically}.\\
While these things are possible by, for example, utilizing the Scala library Scalismo\footnote{https://github.com/unibas-gravis/scalismo}~\cite{luthi2017gaussian}, that approach requires more understanding of the topic, programming effort, and time to set up the development environment. In contrast to that, we concentrated on making this application as easily accessible for everyone as possible and only requiring minimal prior knowledge on the topic in order to be able to use it. The result is a web application that can be accessed by any modern browser without the need to install anything. Our user interface is compatible with the Statismo file format~\cite{luthi2012statismo} and can handle any shape model in that format. For a quick start, we provide a simple face model that has a free license. To the best of our knowledge, this is the first interactive graphical user interface that allows shape model exploration and posterior model creation in the web browser.
\begin{figure}[!t]
    \centering
    \includegraphics[width=\linewidth]{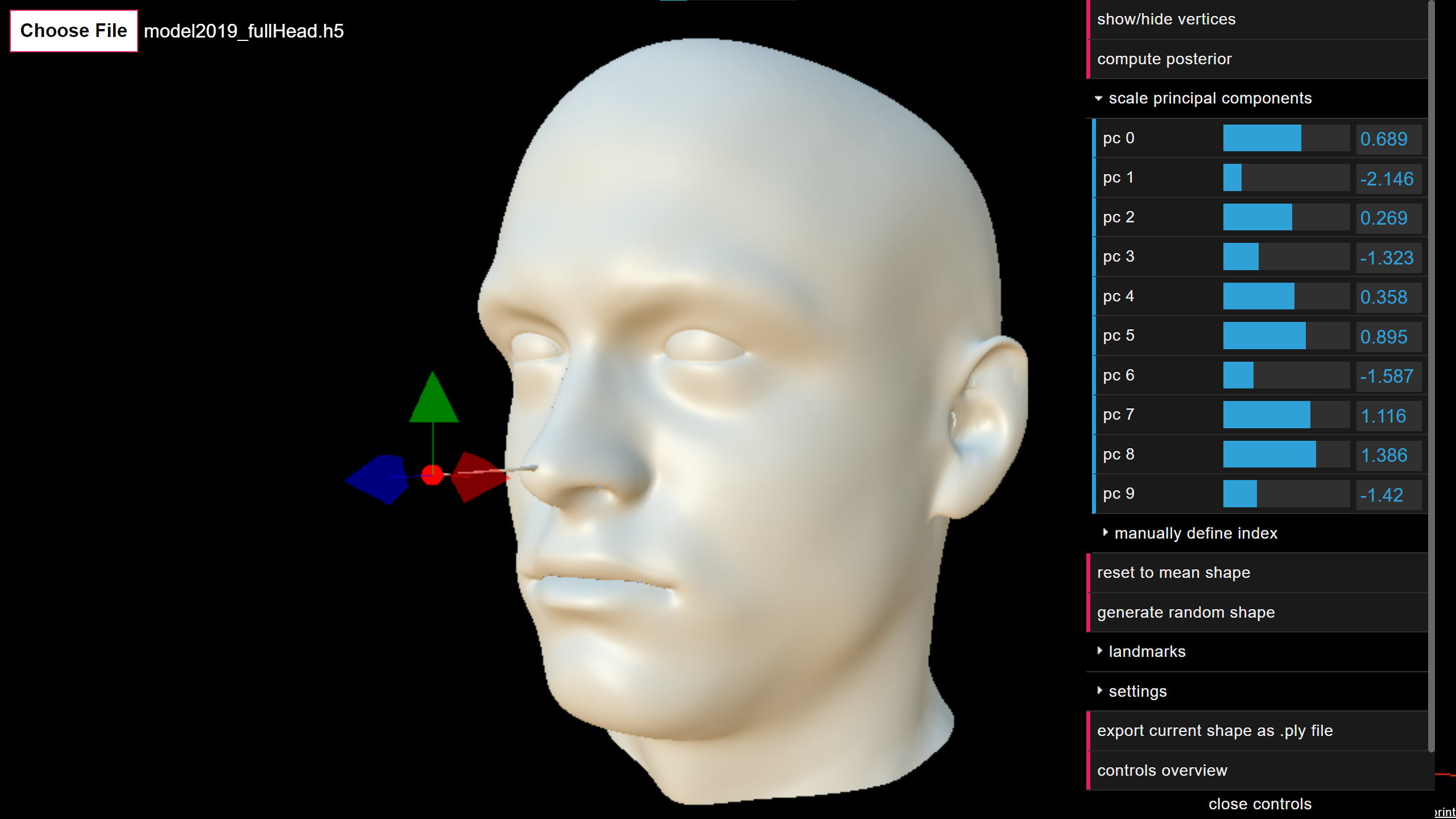}
    \caption{Screenshot of the proposed user interface after generating a random face from the Basel Face Model 2019~\cite{bfm2017paper} and its corresponding principal components on the right. We can now modify the face by adding observations, e.g., here we move the tip of the nose forward and then compute and display the mean face shape of the posterior; size of controls adjusted for readability.}
    \label{fig:user_interface}
\end{figure}
\noindent

\section{Methods}
\textbf{Statistical shape models} are mathematical representations used to capture and analyze the variability of shapes within a given dataset.
SSMs represent this data in a compact way by incorporating a mean shape as well as their variations to that mean shape. \\
A SSM is built from shapes in dense correspondence to a common reference shape allowing to compute the difference in position of each vertex, consisting of an x, y, and z coordinate. Each shape is then represented as a shape vector $s$ containing this difference to the reference shape, the so-called deformation. For this section, we follow the formulation of SSMs as done in~\cite{ albrecht2013posterior}. With this, the mean shape is just the arithmetic mean over all given shape vectors: \(\mu = \frac{1}{n} \sum_{i=1}^n s_i\). To model the class of shapes as a multivariate normal distribution \(\mathcal{N}(\mu,\Sigma)\), the covariance matrix can be computed with \(\Sigma = \frac{1}{n} \sum_{i=1}^n (s_i-\mu) \cdot (s_i-\mu)^T\). Principal component analysis can then be used to decompose the covariance matrix \(\Sigma\) into its eigenvectors and eigenvalues: $\Sigma = U \cdot D^2 \cdot U^T$.
\(U\) is a matrix whose \(i\)-th column is the eigenvector, or principal component, \(q_i\) of \(\Sigma\), ordered from largest to smallest, and \(D^2\) is a diagonal matrix with the corresponding eigenvalues \(D^2_{ii} = \lambda_i\) as diagonal entries. Each principal component now represents an independent characteristic shape variation of the shape class and the eigenvalues quantify the variance \(\sigma^2\), meaning that \(D\) holds the standard deviation \(\sigma\). To generate new shape vectors based on the model, the following equation can be used:
\begin{equation}
    \label{eq:shape_vector}
    s = \mu + Q \cdot \alpha.
\end{equation}
\(\mu\) represents the mean shape, \(Q\) is defined as \(Q = U \cdot D\), holding \(n\) columns with principal components, where each entry of the \(n\) principal components is multiplied with the corresponding standard deviation. \(\alpha\) is then a vector with \(n\) values that follows a standard multivariate normal distribution \(p(\alpha) \sim \mathcal{N}(0,\mathcal{I}_n)\), thus scaling the corresponding principal components after matrix-vector-multiplication. Generating a new normally distributed \(\alpha\) and recalculating the Eq.~\ref{eq:shape_vector} provides a new shape vector \(s\), that represents a new shape variation of the given shape class.
 
\label{sect:posterior}
\noindent
\textbf{Posterior Shape Models: }A particular strength of SSMs is that they are capable of reconstructing a full shape from only partial information. This reconstruction is performed via a posterior shape model, which is a SSM that has been updated by incorporating given observations to calculate a new mean and covariance, the posterior mean and posterior covariance. These observations are the given partial information. They can consist of positional changes of certain vertices specified by the user or of user-created or predefined landmarks of the shape class at hand. \\
To get to the posterior mean and covariance from the given observations, the approach is to repurpose Eq.~\ref{eq:shape_vector} in the sense that the shape vector \(s\) is now given as a partial shape vector \(s_p\), where only those vertices occur that correspond to the observations~\cite{albrecht2013posterior}. Because now \(s_p\) has fewer entries than \(s\), it is essential to also reduce the size of the complete model's \(\mu\) and \(Q\) that appear in Eq.~\ref{eq:shape_vector} for it to be consistent with the dimensions again. Thus, a sub-vector \(\mu_p\) and sub-matrix \(Q_p\) are defined that only include the entries and rows of \(\mu\) and \(Q\), that correspond to the observations in \(s_p\)~\cite{albrecht2013posterior}. This leads to the adjusted equation: $s_p = \mu_p + Q_p \cdot \alpha$.
The next consideration is what values the \(\alpha\) vector has to consist of in order to fulfill the equality. The easiest way to get this \(\alpha\) is to solve the previous equation for it, which leads to:
\begin{equation}
    \label{eq:alpha}
    \alpha = Q_p^{-1} \cdot (s_p - \mu_p).
\end{equation}
Calculating the inverse of \(Q_p\) can be pretty computation-intensive, especially when many observations and principal components are given, as the size of the matrix increases with them. Since \(Q_p\) is also not a quadratic matrix, one has to resort to the generalized or pseudo inverse. There are ways to reduce the complexity and optimize this part of the process~\cite{albrecht2013posterior} but for simplicity reasons this work just relies on the pseudo inverse of \(Q_p\).
The new \(\alpha\) can then be used to compute the posterior mean by just applying it to Eq.~\ref{eq:shape_vector}. The result is a vector that represents the most probable shape in the model factoring in the given observations, the posterior mean.
\begin{figure}[!t]
    \centering
    \includegraphics[width=\linewidth]{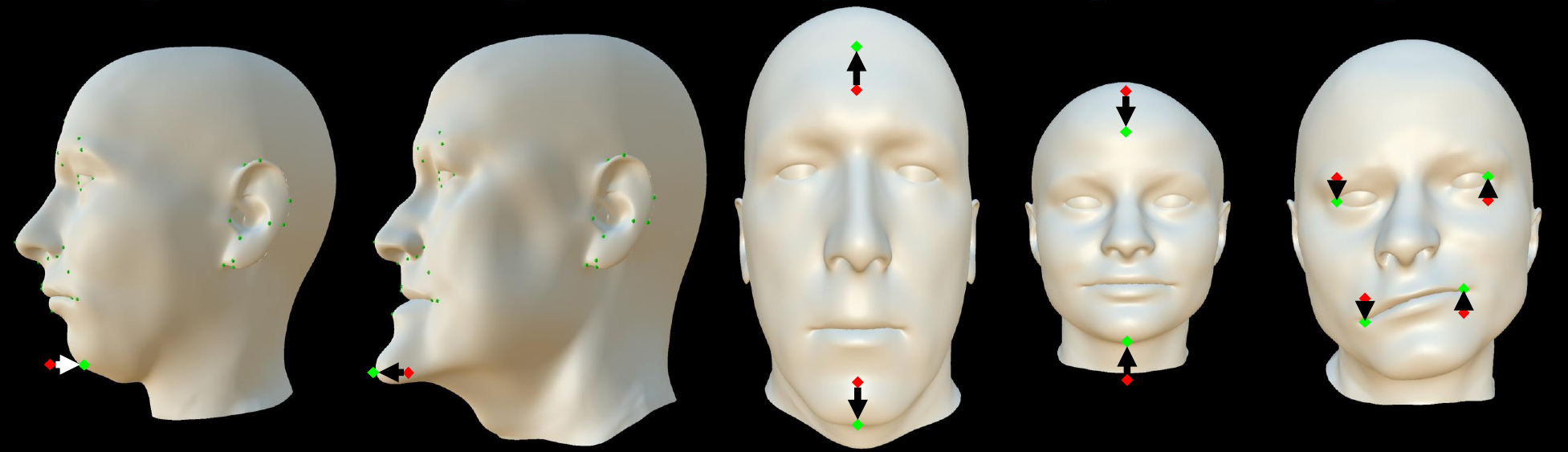}
    \caption{Examples of face manipulations by moving specific points (original position in red, new position in green). The statistical model enables these manipulations and the resulting faces appear natural (from left to right): receding and protruding chin, long face, short face and asymmetric face.}
    \label{fig:posterior_examples}
\end{figure}
\section{Implementation}
Our application is implemented as a web-based application.  We choose JavaScript, HTML and CSS at the core of our implementation, this has the additional benefit that everything runs locally, and no data has to be sent to the server, which is particularly useful when working in a privacy-constrained setting. To realize the 3D component, three.js\footnote{https://github.com/mrdoob/three.js} is used, which is the most popular JavaScript library for displaying 3D content on the web and based on WebGL. With the help of the integrated raycaster and event listeners, we select the nearest vertex of the mesh to the clicked position and mark it with a helper object that represents the 3D coordinate system axes, as can be seen in Figure \ref{fig:user_interface}. With that, the user can move the selected vertex in the desired direction and thus modify the shape. For most of the other controls, dat.GUI\footnote{https://github.com/dataarts/dat.gui} is used as a lightweight controller library that organizes buttons and sliders in a folder-like manner and internally generates the needed HTML and CSS code to display them in the window. It enables interaction with the given 3D scene by changing variables and executing functions. \\
The user can also load in their own SSM if it follows the Statismo file format~\cite{luthi2012statismo}. We extend the implementation using jsfive\footnote{https://github.com/usnistgov/jsfive} to parse the hdf5 files. With the help of the PLYExporter add-on in three.js, the modified face mesh can be exported and downloaded as a .ply file. Additionally, we use the library Toastify.js\footnote{https://github.com/apvarun/toastify-js} for notifications. In order to compute the posterior mean mentioned in section \ref{sect:posterior}, the hardware-accelerated library Tensorflow.js\footnote{https://github.com/tensorflow/tfjs} is used. It provides good performance for matrix operations, which is necessary for large models. Since Tensorflow.js doesn't include a function for the pseudo inverse of a matrix needed for the computation, we had to resort to the worse-performing ml-matrix library\footnote{https://github.com/mljs/matrix} and convert the data types between them to implement Eq.~\ref{eq:alpha}.
\begin{figure}[!t]
    \centering
    \includegraphics[width=\linewidth]{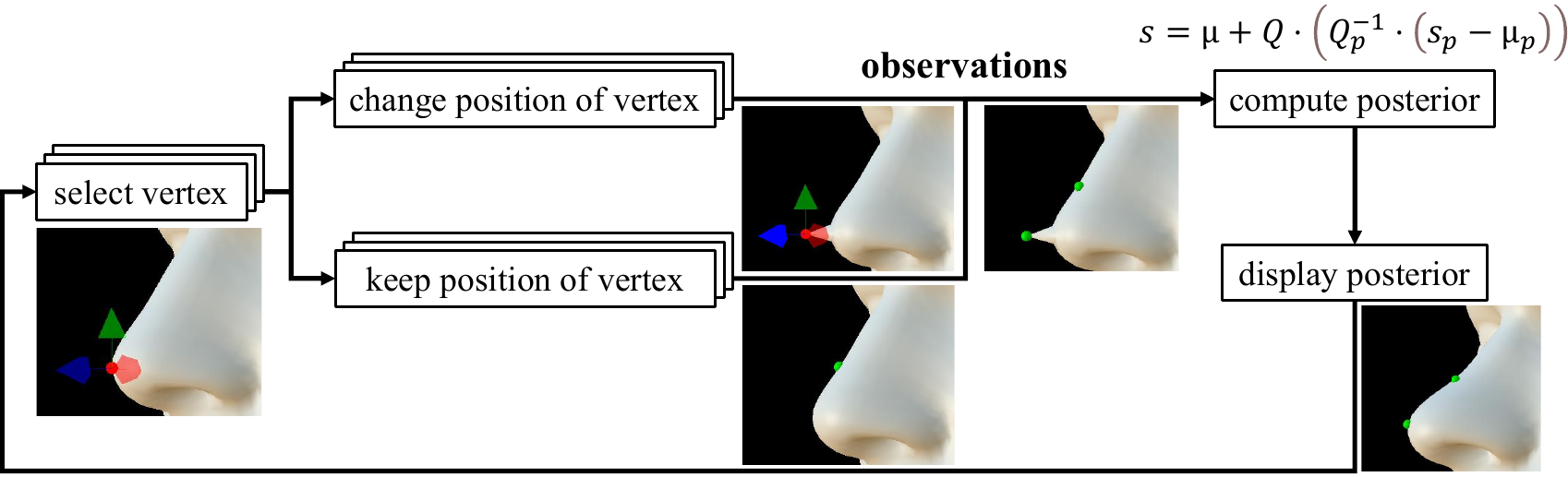}
    \caption{Our interactive design process is structured as follows: we select vertex points on the face surface and decide if they should stay at the current position or if their position should be modified. The posterior is then calculated based on those observations and displayed to the user. The user can then refine the designed face by readjusting existing observations or adding new observations.}
    \label{fig:workflow}
\end{figure}
\section{Results}
\begin{figure} [!t]
    \centering
    \includegraphics[width=\linewidth]{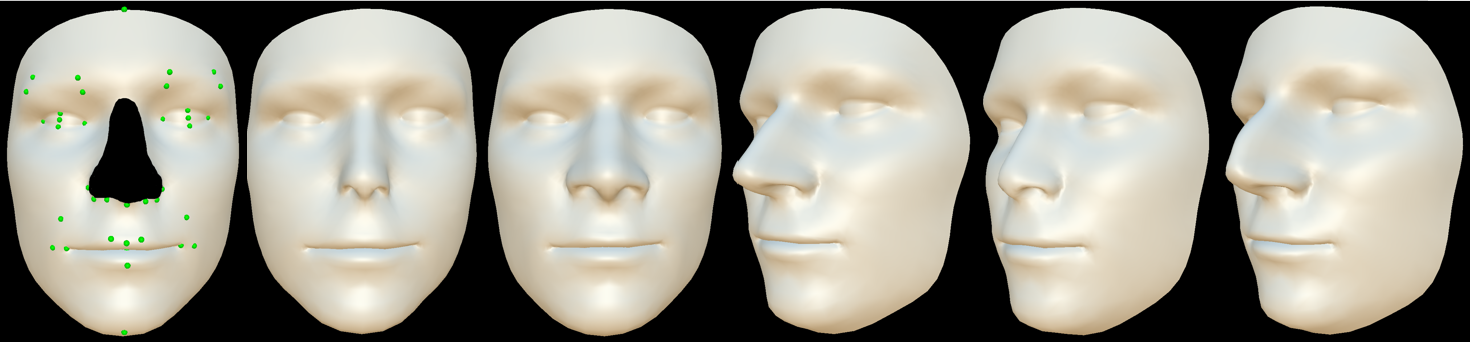}
    \caption{Potential application: We guide different versions for a nose reconstruction for a given face (many observations) and few guiding landmarks on the nose (additional observations) (from left to right): initial shape with missing nose and landmarks, slim, wide, big, small and hooked.}
    \label{fig:nose_reconstruction}
\end{figure}

\begin{figure} [!t]
    \centering
    \includegraphics[width=\linewidth]{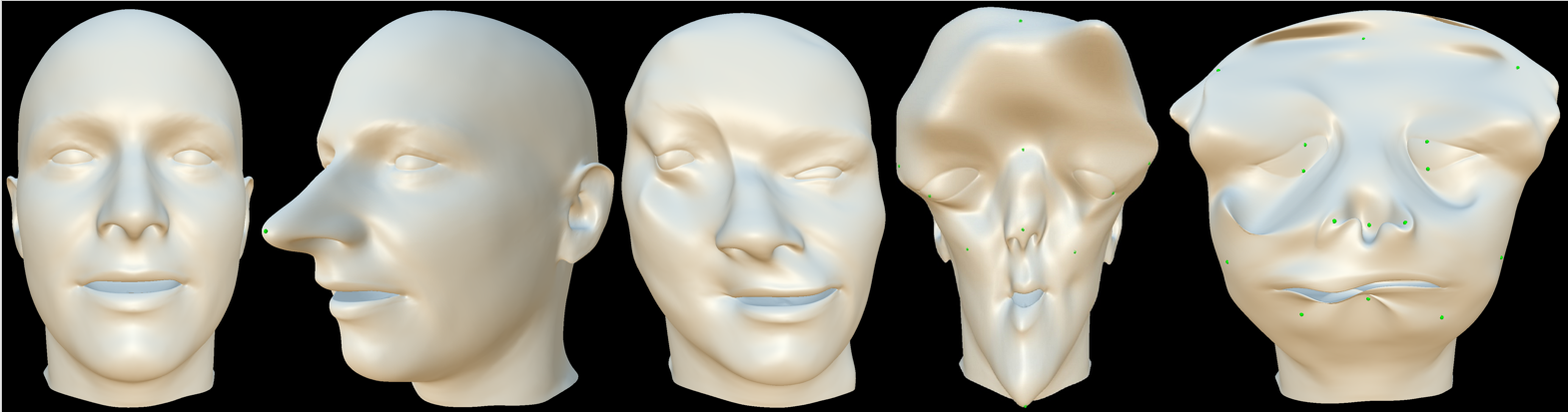}
    \caption{Possible solutions for the tasks in our user study (from left to right): initial shape, shape with long nose, shape with adjusted principal components and two alien-like shapes.}
    \label{fig:tasks}
\end{figure}

\textbf{Creation and Visualization:} We tested our user interface with the Basel Face Model~\cite{bfm2017paper} since it can nicely visualize the features of this application. What the user interface looks like in use can be seen in Figure \ref{fig:user_interface}. Our application allows the user to take all existing principal components of the model into account and generate new shapes by scaling them. They can be defined individually or generated randomly all at once by using the "generate random shape" option and reset to zero by the "reset to mean shape" option.\\
In addition, our implementation offers an easy way to generate characteristic shapes. Figure \ref{fig:posterior_examples} shows such possible shapes for faces.
All these examples were realized by just moving two to four vertices, and were done in less than a minute. How the interactive design process looks like to generate these results can be seen in Figure \ref{fig:workflow}. \\
\textbf{Clinical Use Case: }To demonstrate a potential clinical use case, we present the application of nose reconstruction~\cite{basso2005statistically} in Figure \ref{fig:nose_reconstruction}. We show how different kinds of noses could look like on a given patient's face, with potential application in surgery planning. \\
\textbf{User Study:} In order to determine the usability of our tool, a small user study was performed based on four experts with experience in the field of SSMs. The participants had to complete three tasks with the help of this application:
First, modify the displayed shape by scaling the principal components. Second, modify the shape in a way that it has a very long nose, and third, modify the shape until it looks alien-like to oneself. These tasks were performed on a model based on a single face~\cite{singlescanmodel}. It is a highly flexible model, which is especially helpful when the goal shape doesn't directly relate to a possible human face anymore, such as the face of an alien. Figure \ref{fig:tasks} shows examples of what the resulting shapes could look like.
After completing the tasks, the participants took a small survey. Regarding the difficulty of completing the tasks, the results show a 2.5 on a scale from 1 to 5, the overall experience using the application was a 4 out of 5, and the intuitiveness of the user interface a 3.25. These results are fairly balanced but have a slight tendency toward the desired end of the spectrum. The participants spent 2 to 10 minutes to solve the tasks, being 5 minutes on average, which speaks for how fast one is able to modify shapes in a desired manner using the application. Furthermore, the participants were asked for feedback. One response described the expectation that after deleting all landmarks and recomputing the posterior, the model should reset to its original shape, which it didn't. This behavior was implemented afterward.

\section{Conclusion}
We presented the first web-based GUI to visualize 3D SSMs and explore reconstructions given partial observations. The tool enables fast exploration of basic features of SSMs and can be applied in a teaching setting, to design particular shapes or potentially even in clinical guidance.
The tool is shared as open-source and can be run in the browser.
\begin{acknowledgement}
We thank Timotei Ardelean, Marcel Luthi, Tinashe Mutsvangwa, and Maximilian Weiherer for interesting discussions and feedback. This work was funded by the German Federal Ministry of Education and Research (BMBF), FKZ: 01IS22082 (IRRW). The authors are responsible for the content of this publication.
\end{acknowledgement}
\newpage
\printbibliography

@article{albrecht2013posterior,
    title={Posterior shape models},
    author={Albrecht, Thomas and L{\"u}thi, Marcel and Gerig, Thomas and Vetter, Thomas},
    journal={Medical image analysis},
    volume={17},
    number={8},
    pages={959--973},
    year={2013},
    publisher={Elsevier}
}

@article{luthi2017gaussian,
  title={Gaussian process morphable models},
  author={L{\"u}thi, Marcel and Gerig, Thomas and Jud, Christoph and Vetter, Thomas},
  journal={IEEE transactions on pattern analysis and machine intelligence},
  volume={40},
  number={8},
  pages={1860--1873},
  year={2017},
  publisher={IEEE}
}

@inproceedings{basso2005statistically,
  title={Statistically motivated 3D faces reconstruction},
  author={Basso, Curzio and Vetter, Thomas},
  booktitle={Proceedings of the 2nd international conference on reconstruction of soft facial parts},
  volume={31},
  number={2},
  year={2005},
  organization={Citeseer}
}

@article{luthi2012statismo,
  title={Statismo-A framework for PCA based statistical models},
  author={L{\"u}thi, Marcel and Blanc, Remi and Albrecht, Thomas and Gass, Tobias and Goksel, Orcan and B{\"u}chler, Philippe and Kistler, Michael and Bousleiman, Habib and Reyes, Mauricio and Cattin, Philippe and others},
  journal={The Insight Journal},
  volume={2012},
  pages={1--18},
  year={2012},
  publisher={Kitware, Inc.}
}

@INPROCEEDINGS{bfm2017paper,
  author={Gerig, Thomas and Morel-Forster, Andreas and Blumer, Clemens and Egger, Bernhard and Luthi, Marcel and Schoenborn, Sandro and Vetter, Thomas},
  booktitle={2018 13th IEEE International Conference on Automatic Face \& Gesture Recognition (FG 2018)}, 
  title={Morphable Face Models - An Open Framework}, 
  year={2018},
  volume={},
  number={},
  pages={75-82},
  doi={10.1109/FG.2018.00021}
}

@article{singlescanmodel,
  title={Building 3D Generative Models from Minimal Data},
  author={Sutherland, Skylar and Egger, Bernhard and Tenenbaum, Joshua},
  journal={International Journal of Computer Vision},
  pages={1--26},
  year={2023},
  publisher={Springer}
}

\end{document}